\documentclass{article}
\usepackage{acra}
\usepackage[T1]{fontenc}
\usepackage{amsmath,amssymb}
\usepackage{dsfont}
\usepackage{graphicx}
\usepackage{hyperref}
\usepackage{algorithm}
\usepackage{algpseudocode}
\usepackage{xcolor}
\hypersetup{
  hidelinks,
  pdfauthor={Madeleine Nouri, Gelareh Mohammadi, Francisco Cruz}
}

\usepackage{etoolbox}
\makeatletter
\patchcmd{\@maketitle}
  {\vbox to \titlebox}
  {\vbox}
  {}{}
\patchcmd{\@maketitle}
  {\vskip 0.625in minus 0.125in}
  {\vskip 0pt}
  {}{}
\makeatother

\newcommand{\citeasnote}{%
  \parbox{\textwidth}{\raggedright\small\bfseries
    Cite as: Madeleine Nouri, Gelareh Mohammadi, and Francisco Cruz.
    Demonstration-Free Skill Composition for Navigation and Manipulation.
    \textit{Proceedings of the Australasian Conference on Robotics and
    Automation (ACRA)}, 2026.
  }\\[0.3em]
  \rule{\textwidth}{0.4pt}%
}

\title{%
  \citeasnote\\[0.3em]
  Demonstration-Free Skill Composition for Navigation and Manipulation
}

\hypersetup{
  pdftitle={Demonstration-Free Skill Composition for Navigation and Manipulation}
}

\author{
Madeleine Nouri$^{1}$,
Gelareh Mohammadi$^{1}$,
Francisco Cruz$^{1,2}$\\
$^{1}$School of Computer Science and Engineering,
University of New South Wales, Sydney, Australia\\
$^{2}$Escuela de Ingeniería,
Universidad Central de Chile, Santiago, Chile\\
\{m.nouri, g.mohammadi, f.cruz\}@unsw.edu.au\\
Corresponding author: Madeleine Nouri
}
\begin{document}
\maketitle

\begin{abstract}
Hierarchical reinforcement learning enables sequential tasks to be represented as compositions of separately trained skills. This paper evaluates demonstration-free hierarchical pipelines for MiniGrid navigation and robotic manipulation, comparing their training requirements and task performance with flat reinforcement learning baselines. The evaluation concerns complete pipelines with task-specific observation, reward and control structures. In navigation, two parameterised skills are reused across unseen FourRooms layouts under manually specified high-level plans. In manipulation, reaching, grasp alignment and lifting are trained separately, with downstream skills initialised from states collected during predecessor rollouts. Across unseen navigation layouts, the hierarchical pipeline substantially outperforms the evaluated flat DQN baselines while requiring fewer training interactions. In manipulation, the hierarchical chain requires fewer training steps than flat SAC and exhibits lower observed variation across seeds. The frozen Pick and Lift skills are further reused for stacking by training an additional Place skill. Code and videos are available at \url{https://madeleinenouri.github.io/acra-skill-composition/}.
\end{abstract}
\section{Introduction}
 
Tasks requiring an extended sequence of decisions can be decomposed into a sequence of primitive behaviours. Examples include transitions between rooms in navigation and sequences of reaching, grasping, lifting, transporting and placing in manipulation. Flat reinforcement learning treats the full sequence as a single control
problem, which can lead to poor sample efficiency when exploration must discover a long chain of behaviours from a sparse terminal reward~\cite{hutsebaut-buysseHierarchicalReinforcementLearning2022}.
 
Hierarchical reinforcement learning (HRL) offers a structured alternative in which the task is decomposed into reusable skills, each operating over a shorter horizon with a localised reward signal, and a high-level controller that sequences the skills at execution time~\cite{suttonMDPsSemiMDPsFramework1999,baconOptionCriticArchitecture2016}. Some skill-chaining methods improve composition reliability by
aligning upstream termination distributions with downstream initiation
distributions using expert demonstrations as a reference
\cite{leeAdversarialSkillChaining2021,chenSCaRDualRegularization2024}. Another approach uses engineered rewards instead of demonstrations~\cite{kim2023task,kwon2024reinforcement}. Trained skills can also be reused as components of new
tasks~\cite{hutsebaut-buysseHierarchicalReinforcementLearning2022}. This paper evaluates demonstration-free skill pipelines in navigation
and manipulation, examining training requirements, task performance
and reuse of the trained policies.

The contributions of this paper are:

\begin{enumerate}
    \item An empirical evaluation of demonstration-free hierarchical
    pipelines for discrete navigation and continuous manipulation,
    in which each downstream skill is trained on reset states
    covering its predecessor's success states.
    \item A comparison of pipeline performance and training interactions with the chosen flat baselines, counting cumulative training steps across all skills.
    \item An evaluation of skill reuse on previously unseen navigation layouts without retraining, and extension of manipulation to a stacking task through an additional Place skill.
\end{enumerate}

\section{Related Work}

\subsection{Skill Chaining and Transition Alignment}

Temporal abstraction is commonly formalised through the options framework~\cite{suttonMDPsSemiMDPsFramework1999}, in which each option comprises an initiation set, an intra-option policy, and a termination condition. Option-Critic~\cite{baconOptionCriticArchitecture2016} learns such options end-to-end. Konidaris and Barto~\cite{konidarisSkillChainingReinforcement2009}
introduced skill chaining, training each new skill to reach the initiation
region of the next, and a later extension learns neural initiation classifiers
to select among options~\cite{Bagaria2020Option}. These methods identify
feasible handoff states but do not constrain the termination distribution of
upstream skills.

When independently trained skills fail to compose, a line of work explicitly
aligns their interface. Lee et al.~\cite{leeCOMPOSINGCOMPLEXSKILLS2019} learn a
transition policy that bridges adjacent skills, Byun and
Perrault~\cite{byunTrainingTransitionPolicies2022} cast this as distribution
matching between upstream terminations and downstream initiations, and Chen et
al.~\cite{chenSequentialDexterityChaining2023} chain dexterous manipulation
policies via learned transition controllers. Policy sequencing~\cite{cleggLearningDressSynthesizing2018}
trains each new skill from terminal states produced by the preceding
skill. Adversarial methods instead
regularise the upstream termination distribution directly:
T-STAR~\cite{leeAdversarialSkillChaining2021} uses a GAN-style discriminator
targeting the downstream skill's initiation distribution, and
SCaR~\cite{chenSCaRDualRegularization2024} extends this with bi-directional
regularisation. These adversarial methods depend on expert demonstrations as the
reference distribution for alignment.

\subsection{Reward-Engineered Task Decomposition}
\label{sec:relwork-reward}

The most closely related demonstration-free approaches replace demonstrations with engineered
rewards and independently trained Soft
Actor-Critic~\cite{haarnojaSoftActorCriticOffPolicy2018a} subpolicies. Kim et
al.~\cite{kim2023task} decompose Robosuite pick-and-place into reaching and
grasping subtasks, encoding part of the contact alignment specification in the
upstream reaching reward and executing the grasp through rule-based gripper logic.
Kwon et al.~\cite{kwon2024reinforcement} extend the same template to door
opening, block stacking, and nut assembly.

Our manipulation pipeline belongs to this family, with separate
policies for reaching, grasp alignment and lifting. In Kim et al.'s pick-and-place and Kwon et al.'s block-stacking pipelines, the reaching reward encourages poses suitable for scripted grasping. Our pipeline instead trains grasp alignment as a separate policy and separates lifting from placement, while retaining scripted closure and initial lifting. This decomposition allows us to evaluate extension to stacking with the existing Pick and Lift policies frozen and only Place newly trained.

\section{Method}
\label{sec:method}

This section describes the skill decomposition and execution procedure. The details specific to each environment, including state and action spaces, reward functions and success criteria, are provided in Section~\ref{sec:exp}.

\subsection{Problem Formulation}
\label{sec:skill-decomposition}

A task is a Markov decision process (MDP) $\mathcal{M} = (\mathcal{S}, \mathcal{A},
P, r, \gamma)$ with states $\mathcal{S}$, actions $\mathcal{A}$, transition
kernel $P(s'\mid s,a)$, reward $r$, and discount $\gamma \in [0,1)$. A flat
agent learns a single policy $\pi:\mathcal{S}\rightarrow\mathcal{A}$ maximising
$\mathbb{E}_\pi[\sum_t \gamma^t r(s_t,a_t)]$.

We represent task execution as a sequence of $K$ skill invocations,
$\tau_1 \rightarrow \cdots \rightarrow \tau_K$.
The sequence may invoke the same parameterised policy multiple
times with different task parameters.
Following the options framework~\cite{suttonMDPsSemiMDPsFramework1999}, each skill
is a triple $\tau_i = \langle \mathcal{I}_i, \pi_i, \beta_i \rangle$, where
$\mathcal{I}_i \subseteq \mathcal{S}$ is the \emph{initiation set} (the states
from which $\tau_i$ begins), $\pi_i$ is the intra-skill policy (which may use a
skill-specific observation and action space), and the termination condition is
the indicator of a \emph{success set} $\mathcal{G}_i \subseteq \mathcal{S}$,
$\beta_i(s) = \mathds{1}[s \in \mathcal{G}_i]$. A high-level controller $h$
selects which skill is active. Here $h$ follows a prescribed sequence of skill invocations.
Navigation uses a manually specified plan for each layout, while
Pick and Lift uses a fixed reaching, grasp alignment and lifting
sequence. In general, $h$ may be a learned policy over skills or
a symbolic planner.

A skill may be \emph{parameterised}, $\pi_i(a \mid s, g)$, where $g$ is a task parameter such as a target location or direction, so that a single policy represents multiple related behaviours.  We use parameterised
skills in the discrete domain and unparameterised skills in the continuous
domain.

\subsection{Skill Training and Execution}
\label{sec:ssr}
Reliable execution requires each skill to terminate in states
from which its successor can begin. Following policy
sequencing~\cite{cleggLearningDressSynthesizing2018}, downstream
skills are trained from reset states chosen to cover successful
predecessor handoffs. Reset distributions are specified from task geometry or defined by uniform sampling of successful predecessor terminal simulator states.

We assess composition performance through conditional in-chain
success rates and comparisons with standalone skill performance.

At test time, the high-level controller $h$ executes $\tau_1$ from $s_0 \sim d_0$. A monitor
evaluates the active skill's termination condition, and when $\beta_i(s)=1$
control transfers to $\tau_{i+1}$. The chain succeeds if $\tau_K$ reaches
$\mathcal{G}_K$.

\begin{algorithm}[t]
\caption{Sequential skill training with success-state resets}
\label{alg:ssr}
\begin{algorithmic}[1]
\Require Skills $\tau_1,\dots,\tau_K$, rewards $r_i$, success sets $\mathcal{G}_i$,
  initial distribution $d_0$, rollouts $N$, threshold $\eta$, evaluation episodes $M$
\State $\mu_1 \gets d_0$
\For{$i = 1$ \textbf{to} $K$}
  \State Train $\pi_i$ from $s_0 \sim \mu_i$ until the stopping criterion is met
  \If{$i < K$}
    \State Run $N$ deterministic rollouts of $\pi_i$ from $s_0 \sim \mu_i$
    \State Store the full simulator state at the first successful
       termination of each rollout in pool $\mathcal{B}_i$
    \State $\mu_{i+1} \gets \mathrm{Uniform}(\mathcal{B}_i)$
  \EndIf
\EndFor
\State \Return $\pi_1,\dots,\pi_K$
\end{algorithmic}
\end{algorithm}

\section{Experiments}
\label{sec:exp}
 
We evaluate the approach in two domains: discrete navigation and continuous robotic manipulation. Each subsection specifies its own scenario, MDP, rewards, and training details.
 
\subsection{Experiment 1: Discrete Navigation}
\label{sec:exp-discrete}

\begin{figure}[tbp]
\centering
\includegraphics[width=1\columnwidth]{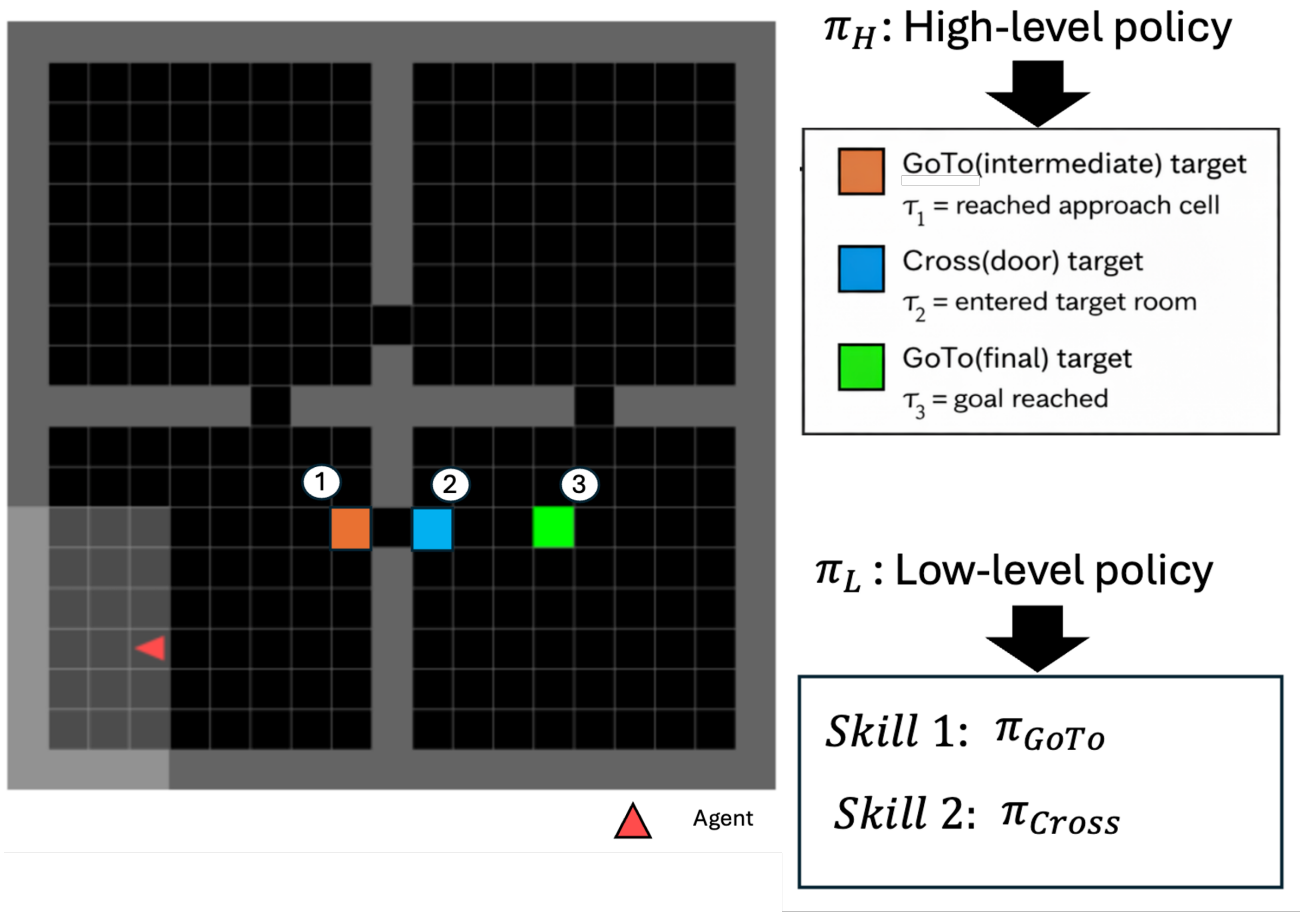}
\caption{MiniGrid FourRooms. The agent (red triangle) navigates from its start
to the goal (green tile), crossing one or more doorways. Door and goal
positions vary across layout seeds. The hierarchical plan is a token sequence
of \textit{GoTo}/\textit{CrossDoor} subtasks ending at the goal.}
\label{fig:minigrid}
\end{figure}

We use the MiniGrid FourRooms
environment~\cite{chevalier-boisvertMinigridMiniworldModular2023} (Fig.~\ref{fig:minigrid}): a
$19\times19$ grid partitioned into four rooms by interior walls, with a single
doorway connecting each adjacent pair of rooms. Each episode starts the agent at
a fixed position. It must reach a goal tile that may lie in another room,
crossing one or more doorways. The action space is restricted to three primitives, \textit{turn left}, \textit{turn right}, \textit{move
forward}. Episodes
terminate on reaching the goal or after $800$ steps. To measure generalisation, we use a pool of $20$ training layouts and $10$
previously unseen evaluation layouts, generated deterministically from fixed global
seeds. Door and goal positions vary across layouts. Each method is trained
over $20$ independent random seeds and evaluated on the $10$ previously unseen layouts
with $50$ episodes per layout per seed.
 
All conditions share the same action space, deterministic transition dynamics
(turn actions rotate $90^\circ$ in place, and move forward advances one
cell unless blocked, in which case the agent remains in place), and discount
$\gamma = 0.99$. Observations and rewards differ per method. Observation definitions and training hyperparameters are given below.

\medskip\noindent\textbf{Doorway-augmented flat DQN.}
A single flat DQN with no decomposition receives a $13$-dimensional observation:
normalised agent pose, goal position, and the coordinates of all four doorways
detected at reset. The flat agent receives the same doorway coordinates as the hierarchical planner. Comparison with the vanilla flat baseline tests the effect of including these coordinates. Reward is
$r = -0.002 + \mathds{1}[\text{goal reached}]$.

 \medskip\noindent\textbf{HRL: \textit{GoTo} subpolicy (parameterised).}
\textit{GoTo} navigates to a target cell (a doorway approach cell or the goal)
within the current room, from an $8$-dimensional observation in \emph{local room
coordinates}: agent pose, clipped displacement to target, and egocentric
free-space indicators. The trained policy is reused across all rooms and previously unseen layouts, with a high-level plan supplied per layout. Reward is
$r = -0.002 + \mathds{1}[\text{target reached}]$.

\medskip\noindent\textbf{HRL: \textit{CrossDoor} subpolicy (parameterised).}
\textit{CrossDoor} crosses a single doorway from the approach cell to the entry
cell, always initialised at the approach cell. Its $8$-dimensional observation appends a one-hot code for the crossing direction to the free-space indicators, allowing a single policy to cover all four directions. Reward is $r = -0.002 + \mathds{1}[\text{entry cell reached}]$.

\medskip\noindent\textbf{HRL: high-level planner.}
The learned subpolicies are sequenced using a manually specified plan for each layout.

\subsubsection{Observation Spaces.}
All observation components are normalised to $[0,1]$. Displacements are affinely rescaled from $[-c,c]$ to $[0,1]$, where $c=6$ cells is the clipping bound. Superscript $n$ denotes a normalised value and subscript $L$ a quantity in local room coordinates. $a_x,a_y$ are the agent's grid position, $d$ its facing direction, and $g_x,g_y$ the goal position. $f_{\text{front}},f_{\text{left}},f_{\text{right}}\in\{0,1\}$ are egocentric free-space indicators, set to 1 if the adjacent cell in that direction is traversable.

\paragraph{Doorway-augmented flat DQN ($13$ dimensions).}
\[
\begin{aligned}
\mathbf{o}=[&a_x^n,a_y^n,d^n,g_x^n,g_y^n,\\
            &d_1^x,d_1^y,\dots,d_4^x,d_4^y],
\end{aligned}
\]
The observation contains the normalised agent pose, goal position and coordinates of the four doorways $d_1,\dots,d_4$ detected at each reset. The flat DQN used in the paired comparison below does not include the eight doorway coordinates, leaving $5$ dimensions.

\paragraph{\textit{GoTo} subpolicy ($8$ dimensions).}
\[
\begin{aligned}
\mathbf{o}=[&a_{xL}^n,a_{yL}^n,\Delta x_L^n,\Delta y_L^n,d^n,\\
            &f_{\text{front}},f_{\text{left}},f_{\text{right}}],
\end{aligned}
\]
where the agent position and the displacement to the target $(\Delta x_L,\Delta y_L)$, clipped to $\pm6$ cells, are expressed in local room coordinates.

\paragraph{\textit{CrossDoor} subpolicy ($8$ dimensions).}
\[
\mathbf{o}=[d^n,f_{\text{front}},f_{\text{left}},f_{\text{right}},\mathbf{e}_{d^*}],
\]
where $\mathbf{e}_{d^*}\in\{0,1\}^4$ one-hot encodes the required crossing direction (right, down, left, or up). \textit{CrossDoor} drops the positional and displacement features of \textit{GoTo}, since it always starts at the doorway approach cell.

\subsubsection{Training Details.}
All policies are trained with Deep Q-Networks
(DQN)~\cite{mnihPlayingAtariDeep2013} via Stable-Baselines3~\cite{stable-baselines3}, chosen for the
discrete action space. The \textit{GoTo} subpolicy is trained on a single fixed
room using
local coordinates and reused across rooms. \textit{CrossDoor} is
trained across the $20$ training layouts to expose it to varied approach configurations.

All discrete-domain policies use the hyperparameters in Table~\ref{tab:dqn_hyperparameters}. \textit{CrossDoor} episodes are capped at $15$ steps. Each skill is trained until its rolling success rate over the last $200$ training episodes reaches $\eta=0.95$, with a minimum of $300$ completed episodes. Rolling training success rates are measured under the exploration policy. All reported results use deterministic evaluation.

\begin{table}[t]
\centering
\begin{tabular}{|l|c|}
\hline
\textbf{Hyperparameter} & \textbf{Value} \\
\hline
Training budget, flat baselines & $500{,}000$ \\
Training budget, \textit{GoTo} & $50{,}000$ \\
Training budget, \textit{CrossDoor} & $50{,}000$ \\
Learning rate & $3\times10^{-4}$ \\
Replay buffer size & $200{,}000$ \\
Learning starts & $5{,}000$ \\
Batch size & $256$ \\
Train frequency & every $4$ steps \\
Discount factor ($\gamma$) & $0.99$ \\
Target network update interval & $2{,}000$ steps \\
Exploration fraction & $0.3$ \\
Final exploration rate & $0.05$ \\
\hline
\end{tabular}
\caption{DQN hyperparameters for the discrete experiment.}
\label{tab:dqn_hyperparameters}
\end{table}

Navigation reset distributions are specified from
room and doorway geometry, without constructing a sampled state pool. \textit{GoTo}’s success condition is satisfied on reaching its commanded target. For a doorway target, its terminal position is the approach cell. \textit{CrossDoor} is initialised there with uniformly random orientation. \textit{GoTo} resets uniformly over the free cells of a room, which contain the entry cells where \textit{CrossDoor} terminates. Approach and entry cells follow from the wall and doorway geometry of each layout.

\begin{figure*}[t]
\centering
\includegraphics[width=\textwidth]{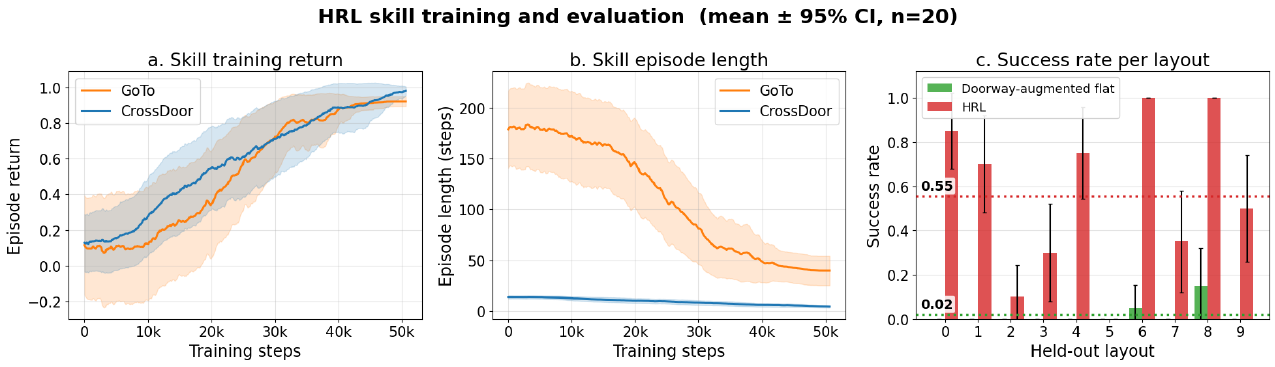}
\caption{Discrete navigation results ($n=20$ seeds, mean, 95\,\% CI).
\textbf{(a)} HRL skill training return and \textbf{(b)} episode length for the
\textit{GoTo} and \textit{CrossDoor} subpolicies. Both subpolicies converge within
$\sim$50{,}000 training steps, with \textit{CrossDoor} converging earlier given its $\sim$5-step horizon.
\textbf{(c)} Success rates across previously unseen layouts for HRL and the flat DQN. Dotted lines mark the overall means ($0.55$ vs $0.02$).}
\label{fig:discrete}
\end{figure*}

\subsubsection{Results.}
Both subpolicies converge within $\sim$50{,}000 steps in return and episode
length (Fig.~\ref{fig:discrete}(a,b)), so skill training costs $\sim$100{,}000
steps in total, whereas the flat baseline, at its full $500{,}000$-step budget, reaches only
$0.020$ on previously unseen layouts (Fig.~\ref{fig:discrete}(c)), HRL achieves a mean success rate of $0.555$
(95\,\% CI $\pm0.069$, 20 seeds, 10 layouts), more than an order of magnitude
above the $0.020$ ($\pm0.019$) of the flat baseline with the same doorway
coordinates. In the reported evaluation rollouts, the flat baseline succeeded only when the goal was in the initial room.

To test the effect of doorway coordinates, we trained a vanilla flat DQN (5-dimensional observation, no door features) under the identical protocol and compared it to the flat DQN with doorway coordinates across $20$ paired seeds. No statistically significant difference was detected in success rate on previously unseen layouts (mean difference $0.010$, 95\,\% CI $\pm0.026$, paired $t$-test $p=0.43$) or asymptotic training return ($p=0.99$).

\subsection{Experiment 2: Continuous Robotic Manipulation}
\label{sec:exp-cont}
 
\begin{figure*}[t]
\centering
\includegraphics[width=0.85\textwidth]{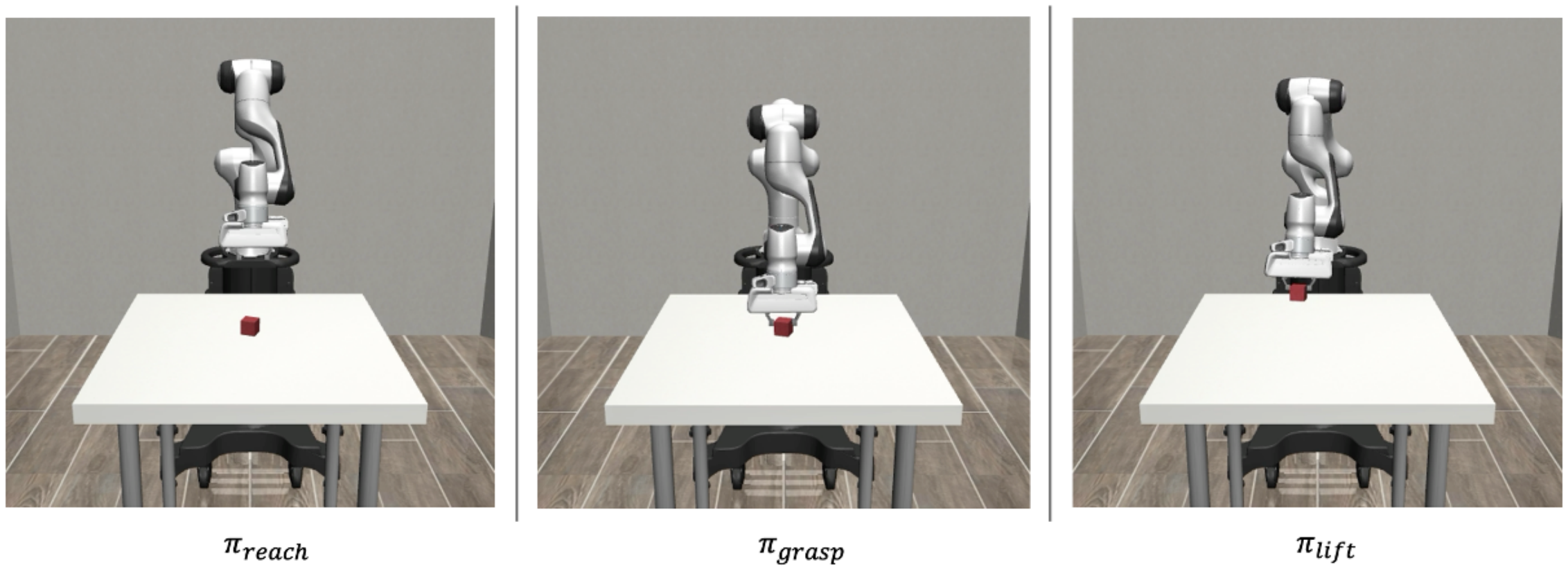}
\caption{Execution of the hierarchical Pick and Lift chain in
Robosuite using a Panda robot arm: left --- approaching the cube
(Reach), centre --- aligning and closing the gripper around the cube
(Grasp), and right --- raising the cube to the target height (Lift).}
\label{fig:robo-env}
\end{figure*}

We use Robosuite~\cite{zhuRobosuiteModularSimulation2025} with a Panda robot arm
(Fig.~\ref{fig:robo-env}). The Pick and Lift task requires the robot to approach a cube,
grasp it with both finger pads and raise it to a target height.
 
The full task is modelled as a Markov decision process
$\mathcal{M} = (\mathcal{S}, \mathcal{A}, P, r, \gamma)$ with continuous state
and action spaces. The state vector contains end-effector position and
orientation, gripper joint positions and velocities, and cube position and
orientation. The action is a 7-dimensional vector specifying end-effector pose
deltas via the Operational Space Controller (OSC) and gripper control.
The transition dynamics are governed by the MuJoCo physics engine with contact-rich
interaction between gripper and cube. The discount factor is $\gamma = 0.99$.

\subsubsection{Hierarchical Decomposition and Transfer.}
The task is decomposed into three sequential skills, each trained as an
independent SAC policy with its own reward, action filter, and termination
condition:
\begin{itemize}
    \item \textbf{Reach:} Move the end-effector toward the cube until the
    gripper--cube distance $d$ falls below $d_{\mathrm{thr}} = 0.04$\,m. The
    action filter disables rotation and locks the gripper open.
    \item \textbf{Grasp:} Align the gripper with the cube and establish contact between both finger pads and the cube. Alignment is measured by two cosine terms,
    $c_{\mathrm{a}}$ between the gripper approach axis and the direction to the
    cube, and $c_{\mathrm{f}}$ between the finger closing axis and the nearest
    cube face normal. Upon meeting the alignment trigger ($d < 0.025$\,m,
    $c_{\mathrm{a}} > 0.5$, $c_{\mathrm{f}} > 0.7$), a rule-based close-and-lift
    sequence executes. Success requires both finger pads in contact with the
    cube and a confirmed lift of $> 5$\,mm.
    \item \textbf{Lift:} Raise the grasped cube to a target height, with a
    curriculum advancing through 3, 5, and 10\,cm. The action filter holds the
    gripper closed.
\end{itemize}
Skills are trained sequentially using Algorithm~\ref{alg:ssr} on full MuJoCo
simulator states.

Each skill is evaluated every $5{,}000$ steps over $M=30$ deterministic episodes. Training stops once its evaluation success rate reaches $\eta=0.95$ on three consecutive evaluations. The reported per-skill step counts correspond to these stopping points. Pools are collected by running $N=500$ deterministic episodes of the trained policy and keeping the complete simulator state when the success condition is triggered. Initial states for downstream skills are sampled uniformly from the pool.

\subsubsection{Reward functions.}
Let $\Delta(x) = \max(0,\, x - x_{\max})$ denote new progress in a quantity $x$,
where $x_{\max}$ is the largest value of $x$ attained so far in the current
episode and is reset at each episode start. Per-skill rewards are:
\begin{align}
r_{\text{reach}} &= -d + \mathds{1}[d < d_{\mathrm{thr}}],\label{eq:reach}  \\
r_{\text{grasp}} &= q\,\Delta(\sigma)
 + b_{\text{align}} \mathds{1}[\text{aligned}]
\notag \\ &\quad
 + b_{\text{grasp}} \mathds{1}[\text{grasp success}], \label{eq:grasp} \\
r_{\text{lift}}  &= \alpha_z \, \Delta(z_{\mathrm{cube}}) \, \mathds{1}[\text{contact}]
\notag \\ &\quad + b_{\text{lift}} \mathds{1}[\text{lift success}],\label{eq:lift}
\end{align}
The alignment predicate $\mathds{1}[\text{aligned}]$ in Eq.~\eqref{eq:grasp} triggers when gripper--cube distance falls below $0.025$\,m with $c_{\text{a}}>0.5$. The rule-based close-and-lift sequence additionally requires $c_{\text{f}}>0.7$. Grasp success requires both finger pads in cube contact, with a confirmed lift of $>5$\,mm validated at least three steps after closure and held for two consecutive steps. The contact predicate in Eq.~\eqref{eq:lift} gates the lift height term on any cube--pad contact.

The pose-shaping score in the grasp reward is the product $\sigma=p(d)\,\omega$, where $p(d)=\mathrm{clip}(1-d/0.08,0,1)$ is the clipped-linear proximity term and $\omega=0.3\,c_{\text{a}}+0.7\,c_{\text{f}}$ is the weighted alignment term. The shaping term rewards only new progress through $\Delta(\sigma)$. The gate $q$ is one until the alignment predicate first triggers and zero thereafter. A per-step penalty of $0.02$ applies throughout the grasp skill.

Per-skill reward coefficients are $\alpha_z=500$, $b_{\text{align}}=1$, $b_{\text{grasp}}=10$, and $b_{\text{lift}}=5$.

For comparison, the flat SAC baseline optimises a shaped reward combining a gripper–cube distance penalty, a reward for new lift progress, and a task-success bonus:
\begin{equation}
r_{\text{flat}} = -\alpha_d \, d + \alpha_p \, \Delta(\text{prog})
 + b_{\text{task}} \mathds{1}[\text{success}],
\label{eq:flat}
\end{equation}
where $\text{prog} \in [0,1]$ is the normalised lift progress. Rewarding only new progress, as in Eq.~\eqref{eq:lift}, prevents repeated reward collection for revisiting an already attained height. The flat baseline therefore receives intermediate shaping, but its reward is not matched to the stage-specific rewards of the hierarchical pipeline. The pipelines also differ in action constraints, rule-based control and replay-buffer handling.

The flat baseline reward coefficients are $\alpha_d=1$, $\alpha_p=10$, and $b_{\text{task}}=5$. The lift reward is awarded only for increases beyond the previously attained maximum height. An earlier absolute height formulation was exploited by hovering at an intermediate height to accumulate reward over successive time steps without completing the task.

\subsubsection{Training Details.}
Both the flat baseline and the hierarchical skills use SAC~\cite{haarnojaSoftActorCriticOffPolicy2018a} with two hidden layers of 256 units each, learning rate $10^{-4}$, batch size 256, discount factor $\gamma=0.99$, target smoothing $\tau=0.002$ ($0.005$ for reach), one gradient step per environment step, and gradient clipping with a maximum norm of $0.5$. Learning starts after $10{,}000$ transitions. The replay buffer capacity is $100{,}000$ transitions for the skill policies and $300{,}000$ for the flat baseline. The entropy coefficient is auto-tuned for reach (default initialisation) and grasp (initialised at $0.1$), and fixed at $0.01$ for lift. The flat baseline uses auto-tuned entropy initialised at $0.1$. The episode horizon is $200$ steps. The grasp policy uses pink action noise~\cite{eberhardPINKNOISEALL2023a} with $\sigma=0.1$, sequence length $200$, and spectral exponent $1.0$. Observations are normalised via running statistics, clipped at $20.0$ and synchronised every $1{,}000$ steps between the training and evaluation environments. Raw actions pass through a skill-specific filter for scaling, deadzone thresholds, exponential moving average smoothing, and gripper overrides.\footnote{After a pool state is loaded into the simulator, the Robosuite OSC controller’s internal goal must be synchronised with the loaded end-effector pose. Otherwise, a one-step transient at episode start can destabilise training.}

The lift skill advances through target heights of 3, 5, and 10\,cm. The target height increases when the rolling success rate over the last $100$ training episodes reaches $0.80$, at which point the replay buffer is flushed to remove transitions whose rewards were computed against the previous height target. The flat baseline uses the same three targets with an advancement threshold of $0.60$ and keeps its replay buffer across stages. All reported lift and full-chain results are at the 10\,cm target.

The flat baseline is a single SAC agent trained end-to-end with $r_{\text{flat}}$
over the full seven-dimensional action space and the same sequence of target heights, for up to
$1{,}000{,}000$ steps. The flat baseline's stopping criterion is assessed at the final $10$\,cm target. The hierarchical skills are trained sequentially with
the reset pools described above (per-skill training steps in Table~\ref{tab:results}). Fifteen seeds
are run per method. The full chain success is evaluated on 30 fixed, unseen
placements sampled from the training distribution (450 rollouts total), and
per-skill success every $5{,}000$ steps over 30 rollouts.

\begin{figure*}[t]
\centering
\includegraphics[width=\textwidth]{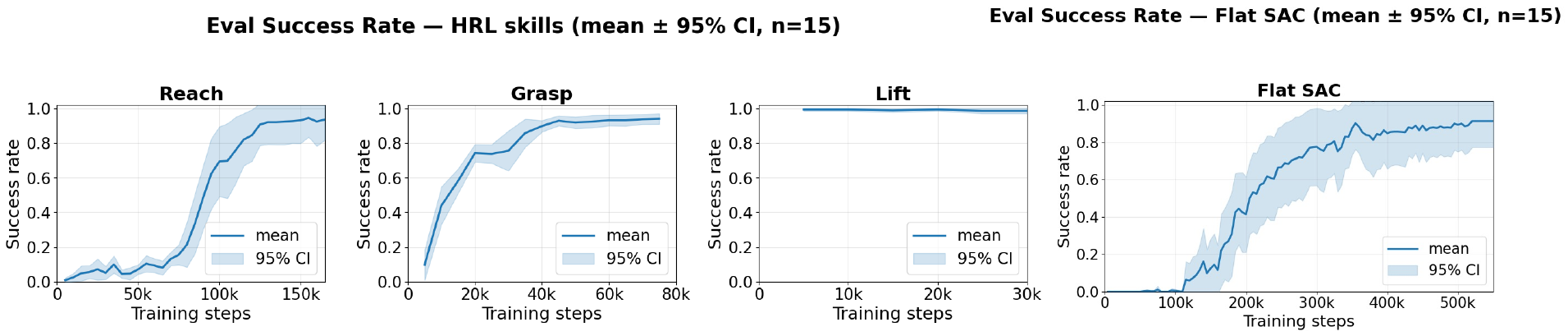}
\caption{Learning curves on the Pick and Lift task ($n=15$ seeds, mean,
95\,\% CI). \textbf{Left:} evaluation success rate versus environment steps for
the three hierarchical skills, converging on average within $\sim$132k (reach),
$\sim$47k (grasp), and $\sim$16k (lift) steps. \textbf{Right:} The flat baseline reaches the training stopping criterion in 14 of 15 seeds ($\sim$292k steps on average). One
seed does not reach the desired success rate within its 1M-step budget.}

\label{fig:hrlcompare}
\end{figure*}
\subsubsection{Results.}
Figure~\ref{fig:hrlcompare} shows learning curves for the
individual skills and the flat baseline. The trained skills form a Pick and Lift policy chain at $\sim$195{,}000
cumulative training steps (Table~\ref{tab:results}). The flat baseline
reaches the stopping criterion in 14 of 15 seeds, requiring $\sim$292{,}000 steps on average
(range 165k--520k). One seed does not reach the desired success rate within its $1{,}000{,}000$-step
budget. The flat baseline's final success rate is $0.914\pm0.253$, compared with $0.858\pm0.103$ for the full chain. The lowest-performing chain seed achieves $0.667$, compared with $0.000$ for the worst flat seed.

At test time the skills are chained by a wrapper that hands off on each skill's
success predicate, with a failure at any skill recorded against that skill. The
chain succeeds in 86\% of rollouts ($0.858\pm0.103$, 95\,\% CI $\pm0.057$, 450
rollouts). Of the 64 failed rollouts, 20 terminated during reaching, 38 during grasping and 6 during lifting.

End-to-end success factorises as the product of per-interface conditional rates
($0.956\times0.912\times0.985\approx0.858$). The pooled conditional success rates differ from the corresponding
mean standalone success rates by less than $0.03$ (Table~\ref{tab:results}),
consistent with limited additional loss at the handoffs. The two training seeds with the lowest full-chain success rates correspond to lower reaching success in one seed ($0.733$ in-chain,
$0.800$ standalone) and lower conditional grasping success in another ($0.700$ conditional), while two trained policies succeeded in all 30 evaluation rollouts.

\begin{table*}[t]
\centering
\begin{tabular}{|l|c|c|c|}
\hline
\textbf{Method} & \textbf{Steps} & \textbf{Success (mean$\pm$std)} & \textbf{95\% CI}\\
\hline
Flat SAC        & $\sim$292k    & $0.914\pm0.253$ & $\pm0.140$\\
\hline
HRL Reach       & $\sim$132k    & $0.978\pm0.051$ & $\pm0.028$\\
HRL Grasp       & $\sim$47k     & $0.940\pm0.055$ & $\pm0.031$\\
HRL Lift (10\,cm) & $\sim$16k   & $0.987\pm0.028$ & $\pm0.015$\\
\hline
\textbf{Full chain} & \textbf{$\sim$195k} & $\mathbf{0.858\pm0.103}$ & $\pm0.057$\\
\hline
\end{tabular}
\caption{Per-skill evaluation success rates (mean\,$\pm$\,std, 95\,\% CI,
15 seeds) and full-chain result over 450 rollouts (30 per seed). Chain
failures: reach 20, grasp 38, lift 6. Flat SAC is reported on the full task. The mean training step count for flat SAC is calculated over the 14 seeds that reached the success criterion, and one seed did not reach this criterion within the budget of $1{,}000{,}000$ steps.}
\label{tab:results}

\end{table*}
 
\subsubsection{Skill reuse in stacking.}
We extend Pick and Lift to stacking by training an additional
Place skill for each of the 15 policy seeds. The Reach,
Grasp and Lift policies and their observation normalisation
statistics remain frozen.

We use a modified Robosuite Stack environment. The upstream
policies retain their original observations of cube~A, without
observing cube~B. Cube~A is sampled within the Lift training
range of $\pm3$\,cm and cube~B within $\pm8$\,cm, subject
to a minimum separation constraint. Cube~A has a fixed
half-width of 2\,cm, compared with 2.0--2.2\,cm during
Lift training.

Place is trained using SAC from successful Lift states
collected in Stack. Its 22-dimensional observation contains
gripper and object states, relative displacements and a
release flag. It controls three translational actions scaled
by $0.3$, with rotation disabled and the gripper closed
until release.

Let $d_{xy}$ be the horizontal distance between cube centres,
$z_A$ the height of cube~A's centre and $z^*$ its target
height when stacked on cube~B. With distances in metres,
the shaping target $z_g(s)$, potential $\Phi(s)$, and reward $r_t$ are
\begin{align}
&z_g(s) = z^* + 0.05\,
\mathrm{clip}\!\left(\frac{d_{xy}-0.005}{0.02},0,1\right),\\
&\Phi(s) = -\frac{d_{xy}+|z_A-z_g(s)|}{0.05},\\
&r_t = q_t[\Phi(s_{t+1})-\Phi(s_t)]
       -0.01 + 10\,\mathds{1}[\text{success}].
\end{align}
where $q_t$ is one before release is triggered and zero
thereafter. Release occurs when $d_{xy}<0.01$\,m and
$|z_A-z^*|<0.015$\,m, followed by a fixed gripper-opening
and upward retreat sequence. A terminal bonus of 10 is awarded when the stacking success condition is satisfied.

Place uses the shared skill-policy SAC settings with
$\tau=0.005$, automatically tuned entropy and the same
pink action noise as Grasp. Training is capped at
$500{,}000$ steps per seed, with evaluation over 30 episodes
every $5{,}000$ steps. Training stops when success reaches
at least $0.95$ on three consecutive evaluations.

During stacking evaluation, the upstream chain has a
600-step budget, with attempt limits of 200, 160 and
250 steps for Reach, Grasp and Lift. On timeout, Reach
continues and Grasp restarts from Reach. Lift restarts
from Reach if the cube has risen by less than 1\,cm
relative to its height at the start of the attempt.
Otherwise, Lift continues.

At 98\% progress towards the 10\,cm Lift target, control
passes to Place for up to 200 steps. Stacking succeeds
when Robosuite's stacking predicate holds for 10 consecutive
control steps after release is triggered.

All 15 policy sets are evaluated deterministically on the
same 30 initialisations, giving 450 episodes. Overall success
is reported as the mean and sample standard deviation
across seeds. Conditional Place success is pooled across
episodes reaching Place.
Table~\ref{tab:stack} reports Pick and Lift completion,
conditional Place success and overall stacking success.

\begin{table}[t]
\centering
\begin{tabular}{lc}
\hline
\textbf{Metric} & \textbf{Result} \\
\hline
Pick and Lift completion & $385/450$ (85.6\%) \\
Place success given handoff & $322/385$ (83.6\%) \\
Full stacking success & $0.716 \pm 0.105$ \\
\hline
\end{tabular}
\caption{Stacking with frozen Pick and Lift policies and an
additionally trained Place skill. Evaluation uses 15 policy
seeds and 30 shared initialisations per seed. Full stacking
success is the mean $\pm$ sample standard deviation across
seeds. The other metrics are pooled episode ratios.}
\label{tab:stack}
\end{table}

Of the 128 unsuccessful episodes, 15 were attributed to
Reach, 40 to Grasp, 10 to Lift and 63 to Place.
Upstream failures are attributed to the first skill whose
handoff condition was never achieved before the chain
budget expired, including retries. Place failures comprise
episodes that completed the Pick and Lift chain but did
not satisfy the stacking success condition.

The Place failure labels comprised 46 timeouts,
16 low-cube states and one displacement of cube~B. Grasp was the largest upstream failure category,
while Place accounted for approximately half of all
unsuccessful episodes.

\section{Discussion}

The navigation results demonstrate reuse of parameterised policies across previously unseen layouts without additional training, with higher success and fewer training steps than the evaluated flat DQN baselines. This transfer uses a manually specified high-level plan for each layout. The flat baseline succeeded only when the goal was located in the initial room, indicating limited performance on tasks requiring inter-room traversal. The addition of doorway coordinates yielded no statistically significant improvement under the evaluated protocol, suggesting that access to these coordinates alone was insufficient to overcome this limitation within the allocated training budget.

In manipulation, the hierarchical chain required fewer training steps than flat SAC and exhibited lower observed variability in final success across seeds, whereas flat SAC achieved a higher mean final success rate. Grasping accounted for the majority of chain failures. Conditional in-chain success rates were close to the corresponding standalone skill success rates, consistent with limited additional performance loss during composition.
The stacking extension achieved $0.716$ success by combining
the frozen Pick and Lift policies with an additionally trained
Place skill. Failures were distributed almost equally between
the upstream chain and placement.

\section{Conclusion}
This paper evaluated demonstration-free hierarchical pipelines for simulated navigation and manipulation. In navigation, two parameterised skills achieved a success rate of $0.555$ on previously unseen FourRooms layouts following approximately $100{,}000$ training steps, compared with $0.020$ for the flat DQN with doorway coordinates after $500{,}000$ steps. In manipulation, the chain achieved $0.858$ success following approximately $195{,}000$ cumulative training steps, compared with $0.914$ for flat SAC, with approximately $292{,}000$ training steps averaged over the 14 of 15 seeds that reached the success criterion. Reusing the frozen Pick and Lift policies with an additionally trained Place skill achieved $0.716$ stacking success.

Several directions remain open. In navigation, these include automated high-level planning, transfer beyond the FourRooms environment and evaluating a flat DQN with the same prescribed subgoals and intermediate rewards as the hierarchical pipeline. In manipulation, future work can include improving skill reliability and investigating flat SAC's higher mean success through additional grasp-specific rewards and ablations of control constraints and training procedures. Learning skill selection, handoff conditions, termination criteria and retry decisions from interaction can reduce reliance on manually specified execution components. Further studies can explore reuse with different predecessors and execution orders through retraining or learned transition policies. Additional comparisons can include reward-engineered and transition-policy methods, alongside execution-time evaluation under matched conditions. A further direction is automatic expansion of the skill set, in which a planner composes existing skills and triggers the training of additional skills to support new transitions.

\section*{Acknowledgments}

This research was supported by the Commonwealth through an Australian Government Research Training Program Scholarship [DOI: \url{https://doi.org/10.82133/C42F-K220}].

\bibliographystyle{named}
\bibliography{references}

\end{document}